\documentclass[11pt]{article}
\usepackage[hyperref]{acl2019}
\usepackage{times}
\usepackage{url}
\usepackage{latexsym}
\usepackage{booktabs}
\usepackage{multirow}
\usepackage{graphicx}  
\usepackage{rotating}
\usepackage{color}
\usepackage{amsmath}

\usepackage{pifont}
\newcommand{\cmark}{\ding{51}}%
\newcommand{\xmark}{\ding{55}}%

\aclfinalcopy 
\def\aclpaperid{1334} 

\title{Know What You Don't Know: Modeling a Pragmatic\\ Speaker that Refers to Objects of Unknown Categories}

\author{Sina Zarrie{\ss} \\
 Faculty of Linguistics and Literary Studies \\
 Bielefeld University, Germany \\
   \texttt{\small sina.zarriess@uni-bielefeld.de} \\\And
      David Schlangen\\
      Linguistics Department\\
       University of Potsdam, Germany\\
        \texttt{\small david.schlangen@uni-potsdam.de} }

\date{}

\begin{document}

\maketitle

\begin{abstract}
 Zero-shot learning in Language \& Vision
is the task of correctly labelling (or \emph{naming}) objects of novel categories. 
Another strand of work in L\&V  aims at pragmatically informative rather than ``correct'' object descriptions, e.g.\ in reference games.
We combine these lines of research and model \textit{zero-shot reference games}, where a speaker needs to successfully refer to a novel object in an image.
Inspired by models of ``rational speech acts'', we extend a neural generator to become a pragmatic speaker reasoning about uncertain object categories. As a result of this reasoning, the generator produces fewer nouns and names of distractor categories as compared to a literal speaker.
We show that this conversational strategy for dealing with novel objects often improves communicative success, in terms of resolution accuracy of an automatic listener.

\end{abstract}

\section{Introduction}


It is commonly agreed that even massive resources for language \& vision~\cite{imagenet,chen2015microsoft,visualgenome} will never fully cover the huge range of objects to be found ``in the wild''. 
This motivates research in zero-shot learning~\cite{lampert2009,socher2013zero,anne2016deep},
which aims at predicting correct labels or names for objects of novel categories, typically via external lexical knowledge such as, e.g., word embeddings.

\begin{figure}
{\footnotesize
\setlength{\tabcolsep}{6pt}
\begin{tabular}{p{3.5cm}p{3.5cm}}
    \raisebox{-\totalheight}{\includegraphics[width = 1\linewidth]{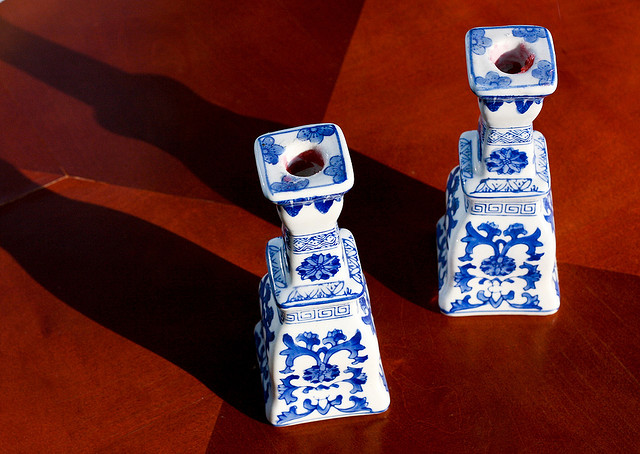}} & 
        \raisebox{-\totalheight}{\includegraphics[width = 1\linewidth]{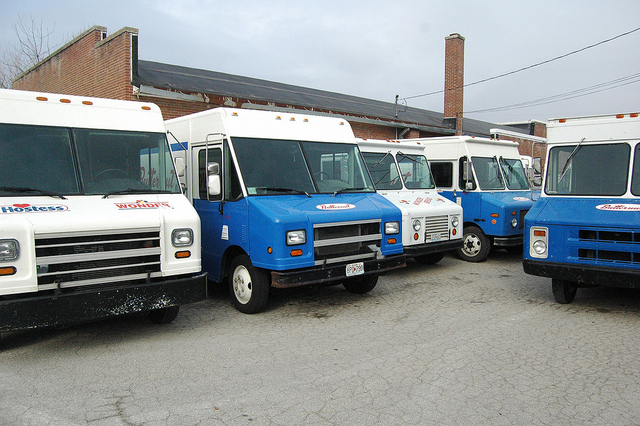}}\\

 \textbf{refexp:} right thingy &   \textbf{refexp:} left blue\\
\end{tabular}
}
\vspace{-0.3cm}
\caption{RefCOCO expressions referring to difficult/unknown objects} 
 \label{fig:im}
 \vspace{-0.3cm}
\end{figure}    

More generally, however, uncertain knowledge of the world that surrounds us, including novel objects, is not only a machine learning challenge: it is simply a very common aspect of human communication, as speakers rarely have perfect representations of their environment. 
Precisely the richness of verbal interaction allows us to communicate these uncertainties and to collaborate towards communicative success~\cite{clarkwilkes:ref}.
Figure~\ref{fig:im} illustrates this general point with two examples from the RefCOCO corpus~\cite{Yu2016}, providing descriptions of visual objects from an interactive reference game.
Here, the use of the unspecific \textit{thingy} and the omission of a noun in \textit{left blue} can be seen as pragmatically plausible strategies that avoid confusing the listener with potentially inaccurate names for difficult-to-name objects.
While there has been a lot of recent and traditional research on pragmatically informative object descriptions in reference games~\cite{mao15,yu2017joint,cohn2018, dale:1995,frank2012predicting}, conversational strategies for dealing with uncertainties like novel categories are largely understudied in computational pragmatics, though see, e.g., work by~\citet{fang2014collaborative}.



 
In this paper, we frame  zero-shot learning as a challenge for pragmatic modeling and explore \textit{zero-shot reference games},
where a speaker needs to describe a novel-category object in an image to an addressee who may or may not know the category.
In contrast to standard reference games, this game explicitly targets a situation where relatively common words like object names are likely to be more inaccurate than other words like e.g.\ attributes.
We hypothesize that Bayesian reasoning in the style of Rational Speech Acts, RSA \cite{frank2012predicting}, can  extend a neural generation model trained to refer to objects of known categories, towards zero-shot learning. 
We implement a Bayesian decoder reasoning about categorical uncertainty and show that, solely as a result of pragmatic decoding, our model produces fewer  misleading object names when being uncertain about the category (just as the speakers did in Figure \ref{fig:im}). Furthermore, we show that this strategy often improves reference resolution accuracies of an automatic listener.

\section{Background}
\label{sec:related}

We investigate referring expression generation (REG henceforth), where the goal is to compute an utterance $u$ that identifies a target referent $r$ among other referents $R$ in a visual scene. 
Research on REG has a long tradition in natural language generation \cite{krahmer:2012}, and has recently been re-discovered in the area of Language \& Vision \cite{mao15,Yu2016,zarriess2018decoding}.
These latter models for REG essentially implement variants of a standard neural image captioning architecture \cite{vinyals:show}, combining a CNN and an LSTM to generate an utterance directly from objects marked via bounding boxes in real-world images.

Our approach combines such a neural REG model with a reasoning component that is inspired by theory-driven Bayesian pragmatics
and RSA \cite{frank2012predicting}. We will briefly sketch this approach here. 
The starting point in RSA is a model of a ``literal speaker'',  $S_0(u|r)$, which generates utterances $u$ for the target $r$. 
The ``pragmatic listener'' $L_0$ then assigns probabilities to all referents $R$ based on the model $S_0$: 
\begin{equation}  
L_0(r|u) \propto \frac{S_0(u|r)*P(r)}{\sum_{r_i \in R}S_0(u|r_i)*P(r_i)}
\label{eq:l0}
\end{equation} 

In turn, the ``pragmatic speaker'' $S_1$ reasons about which utterance is more discriminative and will be resolved to the target by the pragmatic listener:
\begin{equation} 
S_1(u|r) \propto \frac{L_0(r|u)*P(u)}{\sum_{u_i \in U}L_0(r|u_i)*P(u_i)} 
\label{eq:1}
\end{equation} 

\noindent
($S_0$ and $L_0$ are components of the recursive reasoning of $S_1$ and not in fact separate agents.)

There has been some previous work on leveraging RSA-like reasoning for neural language generation.
For instance,  \citet{cohn2018} implement the literal speaker as a neural captioning model trained on non-discriminative image descriptions.
On top of this neural semantics, they build a pragmatic speaker that produces more discriminative captions, applying equation \ref{eq:1} at each step of the inference process.
They evaluate their model in a reference game where an automatic listener (trained on a different portion of the image data) is used to test whether the generated caption singles out the target image among a range of distractor images. A range of related articles have extended neural captioning models with decoding procedures geared towards vocabulary expansion \cite{anderson-guided,nocaps} or contextually discriminative scene descriptions \cite{andreas:2016,vedantam2017context}. 

Previous work on REG commonly looks at visual scenes with multiple referents of identical or similar categories. Here, speakers typically produce expressions composed of a head noun, which names the category of the target, and a set of attributes, which distinguish the target from distractor referents of the same category \cite{krahmer:2012}.
Our work adds an additional dimension of uncertainty to this picture, namely a setting where the category of the target itself might not be known to the model and, hence, cannot be named with reasonable accuracy. 
In this setting, we expect that a literal speaker (e.g.\ a neural REG model trained for a restricted set object categories) generates misleading references, e.g.\ containing incorrect head nouns, as it has no means of ``knowing'' which words risk being inaccurate for referring to novel objects.
The following Section \ref{sec:model} describes how we modify the RSA approach for reasoning in such a zero-shot reference game.

\section{Model}
\label{sec:model}

Inspired by the approach in Section \ref{sec:related}, we model our pragmatic zero-shot speaker as a neural generator (the literal speaker) that is decoded via a pragmatic listener. 
In contrast to the listener in Equation (1), however, our listener possesses an additional latent variable $C$, which reflects its beliefs about the target's category.  
This hidden belief distribution will, in turn, allow the pragmatic speaker to reason about how accurate the words produced by the literal speaker might be.

Our Bayesian listener will assign a probability $P(r|u)$ to a referent $r$ conditioned on the utterance  $u$ by the (literal) speaker.
To do that, it needs to calculate $P(u|r)$, as in Equation \ref{eq:l0}.
While previous work on RSA typically equates $P(u|r)$ with $S_0(u|r)$, we are going to modify the way this probability is calculated.
Thus, we assume that our listener has hidden beliefs about the category of the referent, that we can marginalize over as follows: 
\begin{equation}  
\begin{split}
P(u|r) & = \sum_{c_i \in C} P(u,c_i|r) = \sum_{c_i \in C} \frac{P(u,c_i,r)}{P(r)} \\
  & = \sum_{c_i \in C} \frac{P(r)*P(c_i|r)*P(u|c_i,r)}{P(r)} \\
  & \propto \sum_{c_i \in C} P(c_i|r)*P(u|c_i) 
\end{split}
\end{equation} 



As a simplification, we condition $u$ only on $c_i$, instead of $P(u|c_i,r)$.
This will allow us to estimate $P(u|c_i)$ directly via maximum likelihood on the training data, i.e. in terms of word probabilities conditioned on categories (observed in training) . 
The pragmatic listener is defined as follows:
\begin{equation}  
\begin{split}
L_0(r|u) & = \frac{P(u|r) * P(r)}{P(u)}\\
 & \propto \sum_{c_i \in C} P(c_i|r)*P(u|c_i)
\end{split}
\label{eq:lcat}
\end{equation} 

For instance, let's consider a game with 3 categories and two words, the less specific \textit{left} with $P(u|c_i) = \frac{1}{2}$  for all $c_i \in C$ and the more specific \textit{bus} with  $P(u|c_1) = \frac{9}{10},P(u|c_2) = \frac{1}{10},P(u|c_3) = \frac{1}{10}$. When
the listener is uncertain and predicts $P(c_i|r) = \frac{1}{3}$ for all $c_i \in C$, 
this yields $L_0(r|\text{left}) = 0.5$ and  $L_0(r|\text{bus}) = 0.36$, meaning that the less specific \textit{left} will be more likely resolved to the target $r$. Vice versa, when the listener is more certain, e.g.\ $P(c_1|r) = \frac{9}{10}, P(c_2|r) = \frac{1}{10},P(c_3|r) = \frac{1}{10}$, more specific words will be preferred: $L_0(r|\text{bus}) = 0.83$ and  $L_0(r|\text{left}) = 0.55$.


The definition of the pragmatic speaker is straightforward:  
\begin{equation} 
S_1(u|r)  = S_0(u|r)*L_0(r|u)^{\alpha}
\end{equation}

Intuitively, $S_1$ guides its potentially over-optimistic language model ($S_0$) to be more cautious in producing category-specific words, e.g.\ nouns. 
The idea is that the degree to which a word is category-specific and, hence, risky in a zero-shot reference game can be determined on descriptions for objects of \textit{known} categories and is expressed in $P(u|c)$ .
For \textit{unknown} categories, the pragmatic speaker can deliberately avoid these category-specific words and resort to describing other visual properties like colour or location.\footnote{We leave it for future work to combine this approach with a listener reasoning about  distractor objects in the scene (as in Equation 1).}


Similar to \citet{cohn2018}, we use incremental, word-level inference to decode the pragmatic speaker model in a greedy fashion:
\begin{equation} 
S_1^t(w|r,u_{t-1})  = S_0^t(w|r,u_{t-1})*L_0(r|w)^{\alpha+\beta}
\end{equation} 

At each time step, we generate the most likely word determined via $S_0$ and $L_0$.
The parameters $\alpha$ and $\beta$ will determine the balance between the literal speaker and the listener. While $\alpha$ is simply a constant (set to 2, in our case), $\beta$ is zero as long as $w$ does not occur in $u_{t-1}$ and increases when it does occur in $u_{t-1}$ (it is then set to 2).
This ensures that there is a dynamic tradeoff between the speaker and the listener, i.e.\ for words that occur in previously generated utterance prefix, the language model probabilities ($S_0$) will have comparitively more weight than for new words. 

\section{Exp. 1: Referring without naming?}
\label{subsec:eval1}

Section \ref{sec:model} has introduced a model for referring expression generation (REG) in a zero-shot reference game. 
This model, and its pragmatic decoding component in particular, is designed to avoid words that are specific to categories when there is uncertainty about the category of a target object, in favour of words that are not specific to categories like, e.g., colour or location attributes.
In the following evaluation, we will test how this reasoning component actually affects the referring behavior of the pragmatic speaker as compared to the literal speaker, which we implement as neural supervised REG model along the lines of previous work \cite{mao15,Yu2016}.
As object names typically express category-specific information in referring expressions, we focus the comparison on the nouns generated in the systems' output.

\subsection{Training}

\paragraph{Data} We conduct experiments on RefCOCO \cite{Yu2016} referring expressions to objects in MSCOCO \cite{mscoco} images.
As is commonly done in zero-shot learning, we manually select a range of different categories as targets for our zero-shot game, cf.\ \cite{anne2016deep}.
Out of the 90 categories in MSCOCO, we select 6 medium-frequent categories (\textit{cat,horse,cup,bottle,bus,train}), that are similar to those in \cite{anne2016deep}.
 For each category, we divide the training set of RefCOCO into a new train-test split such that all images with an instance of the target zero-shot category are moved to the test set.

\paragraph{Generation Model ($S_0$)} We implement a standard CNN-LSTM model for REG, trained on pairs of image regions and referring expressions. The architecture follows the baseline version of  \cite{Yu2016}. We crop images to the target region, and obtain the fc features from VGG \cite{vgg}. We set the word embedding layer size to 512, and the hidden state to 1024. We optimized with ADAM, set the batch size to 32 and the learning rate to 0.0004. The number of training epochs is 5 (verified on the RefCOCO validation set).

\paragraph{Uncertainty Estimation}  Similar to previous work in zero-shot learning, we factor out the problem of automatically determining the model's certainty with respect to an object's category, cf. \cite{lampert2009,socher2013zero}: for computing $L_0(r|u)$, we set $P(c_i|r)$ to be a uniform distribution over categories, meaning that the model is maximally uncertain about the referent's category. We leave exploration of a more realistic uncertainty or novelty prediction to future work.

\subsection{Evaluation}

\paragraph{Measures} We test to what extent our models produces incorrect names for novel objects. 
First, for each zero-shot category, we define a set of distractor nouns (\textbf{distr-noun}), which correspond to the names of the remaining categories in MSCOCO.
 Any choice of noun from that set would be wrong, as the categories are pairwise disjunct; the exploration of other nouns (e.g. \textit{thingy, animal}) is left for future work.
In Table \ref{tab:names}, ``\% distr-noun'' refers to how many expressions generated for an instance of a zero-shot category contain such an incorrect distractor noun.
Second, we count how many generated expressions do not contain any noun (\textbf{no-noun}) at all,
according to the NLTK POS tagger.

\paragraph{Results}


\begin{table}
\centering 
\small
\begin{tabular}{llccc}
\toprule
 \multicolumn{2}{c}{Model} & \% distr-noun & \% no-noun\\
\midrule
\textit{cat} & $S_0$ & 0.606 & 0.107\\
& $S_1$  & \bf 0.484 & \bf 0.193\\
\midrule
\textit{horse} & $S_0$ & 0.683 & 0.085\\
 & $S_1$  & \bf 0.572 &\bf  0.30\\
\midrule
\textit{cup} & $S_0$ & 0.627  & 0.079\\
 & $S_1$ & \bf 0.332 & \bf 0.172\\
  \midrule
  \textit{bottle} & $S_0$ & 0.398 & 0.275\\
   & $S_1$  & \bf 0.166 & \bf 0.562\\
   \midrule
  \textit{bus} & $S_0$ & 0.743 & 0.066\\
    & $S_1$  & \bf 0.612 & \bf 0.247\\
  \midrule
  \textit{train} & $S_0$ & 0.759 & 0.166\\
  & $S_1$  & \bf 0.558 & \bf 0.37\\
\bottomrule
\end{tabular}
\caption{Names and nouns contained in generation output  for two speakers ($S_0, S_1$)}
\label{tab:names}
\end{table}

Table \ref{tab:names} shows that the proportion of output expressions containing a distractor noun \textit{decreases} markedly from $S_0$ to $S_1$, whereas the proportion of expression without any name \textit{increases} markedly from $S_0$ to $S_1$. 
First of all, this suggests that our baseline model $S_0$ does, in many cases, not know what it does not know, i.e.\ it is not aware that it encounters a novel category and frequently generates names of known categories encountered during training.
However, even in this simple model, we find a certain portion of output expressions that do not contain any name (e.g.\ 27\% for \textit{bottle}, but only 6\% for \textit{bus}).
The results also confirm our hypothesis that the pragmatic speaker $S_1$ avoids to produce ``risky'' or specific words that are likely to be confused for uncertain or unknown categories.
It is worth stressing here that this behaviour results entirely from the Bayesian reasoning that $S_1$ uses in decoding; the model does not have explicit knowledge of linguistic categories like nouns, names or other taxonomic knowledge.

\section{Exp. 2: Communicative success}
\label{subsec:eval2}

\begin{figure}
{\footnotesize
\setlength{\tabcolsep}{6pt}
\begin{tabular}{p{4cm}p{3.5cm}}

 \raisebox{-\totalheight}{\includegraphics[width = 1\linewidth]{COCO_train2014_000000317210.jpg}}  & \vspace{0.2cm} \shortstack{Target (unknown cat):\\ left horse\\ \vspace{0.2cm} \\\textbf{$S_0$:} left person \xmark \\ \\
   \textbf{$S_1$:} left black  \cmark} \\
\end{tabular}
}
\vspace{-0.3cm}
\caption{Qualitative Example} 
 \label{fig:im2}
 \vspace{-0.3cm}
\end{figure}    

The Experiment in Section \ref{subsec:eval1} found that the pragmatic speaker uses less category-specific vocabulary when referring to objects of novel categories as compared to a literal speaker.
Now, we need to establish whether the resulting utterances still achieve communicative success in the zero-shot reference game, despite using less specific vocabulary (as shown above).
We test this automatically using a model of a ``competent'' listener, that knows the respective object categories. This is supposed to approximate  a conversation between a system and a human that has more elaborate knowledge of the world than the system. 

\begin{table}
\begin{tabular}{lc}
\toprule
 Zero-shot category & Similar category\\
\midrule
\textit{cat} & dog, cow \\
\textit{horse} & dog, cow \\
\textit{cup} & bowl, bottle, wine glass\\
  \textit{bottle} & vase, wine glass\\
  \textit{bus} & car, train, truck \\
  \textit{train} & car, bus, truck\\
 \bottomrule
\end{tabular}
\caption{Target and distractor categories used for testing in Exp. 2}
\label{tab:evalcat}
\end{table}

\paragraph{The evaluation listener}
One pitfall of using a trained listener model (instead of a human) for task-oriented evaluation is that this model might simply make the same mistakes as the speaker model as it is trained on similar data.
To avoid this circularity, \citet{cohn2018} train their listener on a different subset of the image data.
Rather than training on \textit{different} data, we opt for training the listener on \textit{better} data, as we want it to be as strict and human-like as possible. 
For instance, we do not want our listener model to resolve an expression like \textit{the brown cat} to a \textit{dog}.
We train $S_{eval}$ as a neural speaker on the entire training set and give $L_{eval}$ access to ground-truth object categories.
The ground-truth category $c_r$ of a referent $r$ is used to calculate $P(n_u|c_r)$ where $n_u$ is the object name contained in the utterance $u$. $P(n_u|c_r)$  is estimated on the entire training set.
\begin{equation}  
L_{eval}(r|u,c_r) = S_{eval}(u|r) * P(n_u|c_r)
\end{equation} 

 $P(n_u|c_r)$  will be close to zero if the utterance contains a rare or wrong name for the category $c_r$, and $L_{eval}$ will then assign a very low probability to this referent. 
We apply this listener to all referents in the scene and take the argmax.

\paragraph{Test set}

The set \textbf{TS-image} pairs each target with other (annotated!) objects in the same image, a typical set-up for reference resolution.
As many images in RefCOCO only have distractors of the \textit{same} category as the target (which is not ideal for our purposes), we randomly sample an additional test set called \textbf{TS-distractors}, pairing zero-shot targets with 4 distractors of a similar category, which we defined manually, shown in Table \ref{tab:evalcat}.  
This is slightly artificial as objects are taken out of the coherent spatial context, but it helps us determining whether our model can successfully refer in a context with similar, but not identical, categories.

\paragraph{Results}

\begin{table}
\centering
\small
\begin{tabular}{llcc}
\toprule
 \multicolumn{2}{c}{Model}   & TS-image & TS-distractors\\
\midrule
\multirow{ 2}{*}{\textit{cat}} & $S_0$ & 0.516 & 0.343\\
   & $S_1$ & \bf 0.603 & \bf  0.386\\
\midrule
\multirow{ 2}{*}{\textit{horse}} & $S_0$ & \bf 0.644 & 0.096\\
   & $S_1$ & 0.589 &  \bf 0.150\\
  \midrule
\multirow{ 2}{*}{\textit{cup}} & $S_0$ & \bf 0.721 & 0.483\\
   & $S_1$ & 0.674 &  \bf 0.540\\
    \midrule
\multirow{ 2}{*}{\textit{bottle}} & $S_0$ & 0.502 & 0.275\\
   & $S_1$  & \bf 0.517 & \bf 0.306\\
   \midrule
   \multirow{ 2}{*}{\textit{bus}} & $S_0$ & \bf 0.789 & \bf 0.405\\
   & $S_1$  & 0.759 & 0.361\\
    \midrule
   \multirow{ 2}{*}{\textit{train}} & $S_0$ & 0.658 & 0.202\\
   & $S_1$  & \bf 0.667 & \bf 0.305\\
\bottomrule
\end{tabular}
\caption{Reference resolution accuracies obtained from listener $L_{eval}$ on expressions by $S_0, S_1$}
\label{tab:listener}
\vspace{-0.5cm}
\end{table}

As shown in Table \ref{tab:listener}, the $S_1$ model improves the resolution accuracy for all categories on TS-distractors, except for \textit{bus}.
On TS-image, resolution accuracies are generally much higher and the comparison between $S_0$ and $S_1$ gives mixed results.
We take this as positive evidence that $S_1$ improves communicative success in a relevant number of cases, but it also indicates that combining this model with the more standard RSA approach could be promising. Figure \ref{fig:im2} shows a qualitative example for $S_1$ being more successful than $S_0$. 

\section{Conclusion}

We have presented a pragmatic approach to modeling zero-shot reference games, showing that Bayesian reasoning inspired by RSA can help decoding a neural generator that refers to novel objects. The decoder is based on a pragmatic listener that has hidden beliefs about a referent's category, which leads the pragmatic speaker to use fewer nouns when being uncertain about this category. 
While some aspects of the experimental setting are, admittedly, simplified (e.g. compilation of an artificial test set, uncertainty estimation), we believe that this is an encouraging result for scaling models in computational pragmatics to real-world conversation and its complexities.


%
%

\bibliography{conll2018}
\bibliographystyle{acl_natbib}

\end{document}